\documentclass[letterpaper, 10 pt, conference]{ieeeconf}  

\IEEEoverridecommandlockouts                              

\usepackage{graphics} 
\usepackage{graphicx}
\usepackage{pifont}
\newcommand{\cmark}{\ding{51}}%
\newcommand{\xmark}{\ding{55}}%
\usepackage{verbatim}
\usepackage{color,soul}
\usepackage{multirow}
\usepackage{amsmath} 
\usepackage{dsfont}
\usepackage{tabu}
\usepackage{xcolor}
\usepackage{colortbl}
\usepackage[font=small]{caption}
\usepackage{subcaption}
\usepackage[%
    colorlinks=true,
    pdfborder={0 0 0},
    linkcolor=red
]{hyperref}
\usepackage[ruled, linesnumbered, lined]{algorithm2e}
\title{\LARGE \bf
Suction Leap-Hand: Suction Cups on a Multi-fingered Hand Enable Embodied Dexterity and In-Hand Teleoperation
}

\author{Sun Zhaole$^{1}$, Xiaofeng Mao$^{1}$, Jihong Zhu$^{2}$, Yuanlong Zhang$^{3}$,  Robert B. Fisher$^{1}$
\thanks{$^{1}$Sun Zhaole, Xiaofeng Mao, and Robert B. Fisher are with the School of Informatics, University of Edinburgh, UK.
Contact: 
        {\tt\small szlhszt@outlook.com, zhaole.sun@ed.ac.uk}
        }%
\thanks{$^{2}$Jihong Zhu is with School of Physics, Engineering and Technology, University of York, UK}
\thanks{$^{3}$Yuanlong Zhang is with Tsinghua University, China}
}

\begin{document}

\maketitle
\thispagestyle{empty}
\pagestyle{empty}

\begin{abstract}
Dexterous in-hand manipulation remains a foundational challenge in robotics, with progress often constrained by the prevailing paradigm of imitating the human hand.
This anthropomorphic approach creates two critical barriers: 1) it limits robotic capabilities to tasks humans can already perform, and 2) it makes data collection for learning-based methods exceedingly difficult.
Both challenges are caused by traditional force-closure which requires coordinating complex, multi-point contacts based on friction, normal force, and gravity to grasp an object. 
This makes teleoperated demonstrations unstable and amplifies the sim-to-real gap for reinforcement learning. 
In this work, we propose a paradigm shift: moving away from replicating human mechanics toward the design of novel robotic embodiments.
We introduce the \textbf{S}uction \textbf{Leap}-Hand (SLeap Hand), a multi-fingered hand featuring integrated fingertip suction cups that realize a new form of suction-enabled dexterity.
By replacing complex force-closure grasps with stable, single-point adhesion, our design fundamentally simplifies in-hand teleoperation and facilitates the collection of high-quality demonstration data.
More importantly, this suction-based embodiment unlocks a new class of dexterous skills that are difficult or even impossible for the human hand, such as one-handed paper cutting and in-hand writing. 
Our work demonstrates that by moving beyond anthropomorphic constraints, novel embodiments can not only lower the barrier for collecting robust manipulation data but also enable the stable, single-handed completion of tasks that would typically require two human hands.
Our webpage is \url{https://sites.google.com/view/sleaphand}.
\end{abstract}

\section{Introduction}
\label{sec:intro}
Dexterous manipulation, the ability to reconfigure objects within a single hand, remains a grand challenge in robotics \cite{billard2019trends, bicchi2002hands}. 
The dominant paradigm for achieving this goal has been data-driven learning on anthropomorphic hands, an approach that has led to successes in grasping and reorientation \cite{chen2023visual, akkaya2019solving, wang2023dexgraspnet}. 
However, this focus on human-like forms has also fundamentally tethered robotic potential to the limits of human dexterity. 
Current research is largely centered on replicating those simple tasks that human hands can do, rather than solving challenging dexterous tasks or exploring what novel robotic morphologies might achieve. 
This reality raises our core question: to achieve advanced dexterity, must robots merely imitate the human hand, or can they leap beyond these biological constraints by exploiting a unique physical embodiment?

Recent work has begun to challenge this human-centric paradigm by demonstrating that dexterity emerges not only from sophisticated control but also from morphology itself \cite{gilday2025embodied}. A systematic exploration of both anthropomorphic and non-anthropomorphic designs revealed that novel configurations can yield unique, emergent manipulation behaviors without complex controllers. This provides compelling evidence that robotic dexterity is not intrinsically bound to the imitation of the human form. 

\begin{figure}[t]
  \centering
  \includegraphics[width=0.48\textwidth]{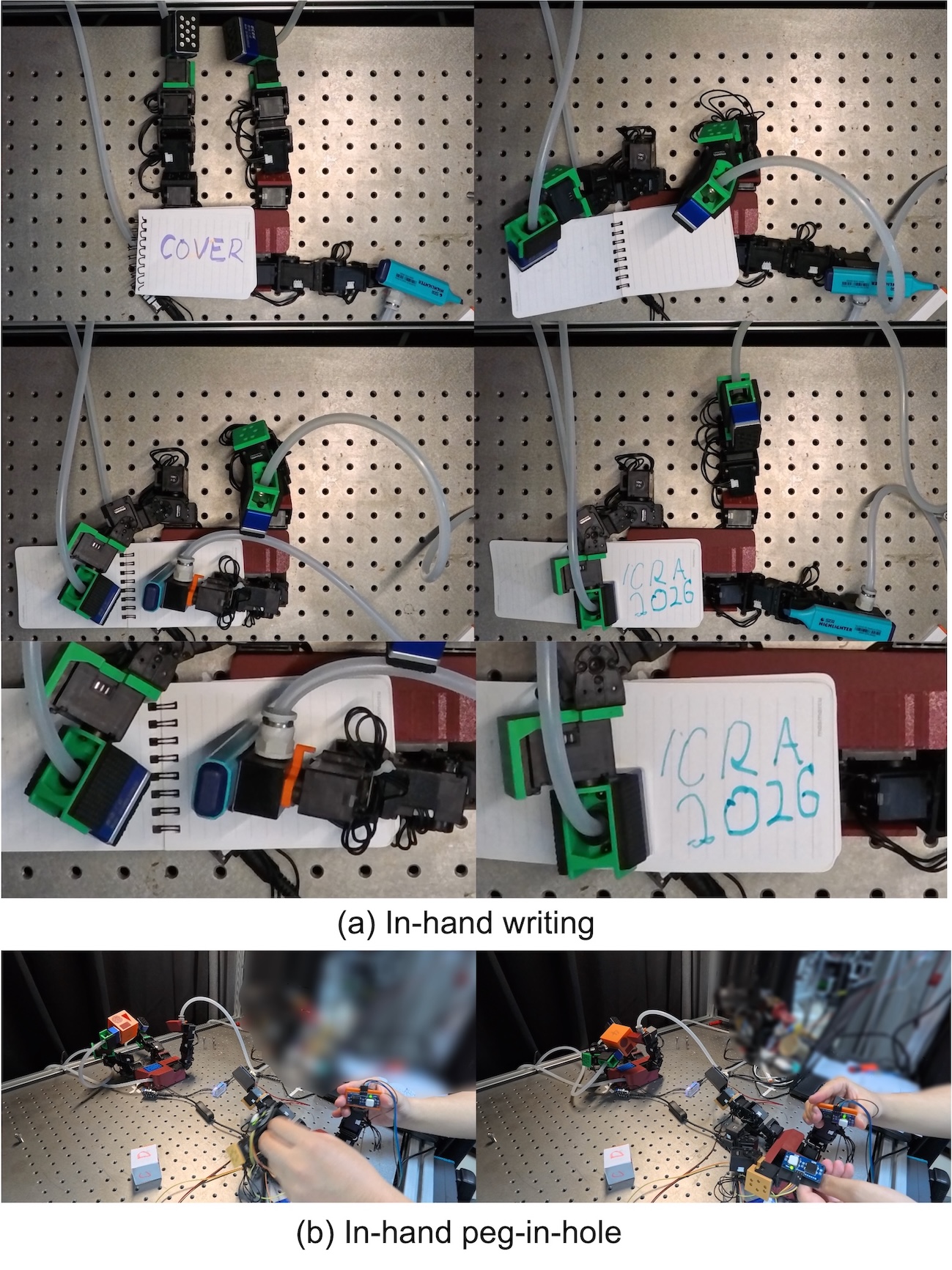}
  \caption{Teleoperating the SLeap Hand performing challenging in-hand manipulation relying on suction-based embodiment. a) In-hand writing. b) In-hand peg-in-hole with the scene of teleoperations.}
  \label{fig:intro_demo}
  \vspace{-7mm}
\end{figure}
 
Despite this insight, the current anthropomorphic approach has created two interconnected barriers that impede progress. 
Both barriers are caused by a deep reliance on force-closure to maintain a stable grasp through a balance of gravity, normal forces, and friction across multiple contact points.

The first barrier is the extreme difficulty of teleoperating dexterous in-hand manipulation. 
Whether using wearable exoskeletons \cite{wei2024wearable, zhang2025doglove} or motion capture systems \cite{arunachalam2022dexterous, tao2025dexwild}, the human operator is cognitively burdened with constantly managing multiple contact points to prevent grasp failure during long-horizon tasks.
This makes the collection of high-quality, successful demonstrations inefficient and unreliable. 
The second barrier is the inherent limitation of the anthropomorphic designs themselves. 
Designs like the Shadow Hand and Allegro Hand are mechanically anthropomorphic, but are also bounded by the functions and dexterity as the human hand. 
Those complex dexterous manipulations are still within human hands' capabilities, e.g., in-hand arbitrary object reorientation and solving Rubik's Cube \cite{chen2023visual, akkaya2019solving}.
The root cause of both challenges is a reliance on force-closure: the need to form a stable grasp through a precise balance of gravity, normal forces, and friction across multiple contact points.
As a result, even the simplest primitive, grasping, typically requires at least two fingers or a finger and the palm.
For imitation learning, this makes it incredibly difficult for a human to demonstrate complex in-hand motions without dropping the object, resulting in easily failed demonstrations in long-horizon tasks.
For reinforcement learning, this reliance on rich, sensitive contact dynamics massively complicates sim-to-real transfer, requiring much effort for real-world deployment. 
By continuing to follow this principle, we are not only making data collection harder but also capping robotic dexterity at the human level. 

To overcome these limitations, we propose a paradigm shift from friction-based force-closure to adhesion-based manipulation. 
We introduce the \textbf{S}uction \textbf{Leap}-Hand (SLeap Hand), a 5-DoF-per-finger robotic hand where each fingertip is augmented with a suction cup, shown in Figure \ref{fig:intro_demo}.
This suction-based embodiment directly addresses the two core challenges.
First, it dramatically simplifies teleoperation. 
A single suctioned fingertip can create a firm grasp, freeing the operator to focus on the manipulation task itself without the constant risk of dropping the object.
This lowers the barrier to collecting high-quality data for imitation learning. 
The teleoperation can be paused at any time to reduce operators' workload without keeping lifting arms in the air or wearing exoskeletons all the time.
Second, and more importantly, it unlocks new dexterous skills. 
By leveraging adhesion, the SLeap Hand can perform tasks that are exceedingly difficult or impossible for a single human hand, such as cutting paper or writing on a notebook held by the same hand.
  
This work shows that by exploring suction-based embodiments that extend beyond human hand capabilities, the SLeap Hand enables teleoperation of complex dexterous tasks.
The main contributions of this paper are:

\begin{itemize}
    
    \item A novel hardware embodiment: 
    We present the design and realization of the SLeap Hand, a 15-DoF, three-fingered manipulator featuring suction cups on each fingertip and the palm. 
    This design facilitates a hybrid manipulation strategy, combining traditional frictional contacts with controlled adhesion.

    \item An teleoperation system to collect challenging dexterous demonstrations precisely and comfortably: 
    We develop a teleoperation system that uniquely leverages the hand's suction capabilities to provide stable and intuitive control. 
    This enables human to generate high-fidelity demonstrations of complex tasks in a quasi-static way without the assistance of learned models or complex controllers on non-anthropomorphic hands.
    Besides, it reduces operators' workload.

    \item Demonstration of advanced dexterity: 
    We conduct a series of demonstrations on challenging manipulation tasks, such as paper cutting and in-hand writing, to benchmark the system's performance. 
    The results demonstrate that our SLeap Hand with suction-based embodiment enables the next level of dexterity previously difficult to achieve with conventional robotic hands.

\end{itemize}

\section{Related Work}
\label{sec:related_work}

\textbf{Dexterous manipulation} is about changing objects' poses by the palm and fingers within the hand space \cite{mason2018toward, bullock2012hand, bicchi2002hands}. 
Common dexterous manipulation tasks include object grasping \cite{wang2023dexgraspnet}, reorientation \cite{chen2022system, chen2023visual}, solving a Rubik's cube \cite{akkaya2019solving}, multiple object grasping \cite{yao2023exploiting} and so on. 
Existing demonstrations of robotic dexterity typically require significant investment in simulation or task-specific learning, as direct teleoperation is often too difficult for collecting high-quality demonstrations.
Moreover, these demonstrated tasks, even highly complex ones like in-hand Rubik's cube solving, are fundamentally bound by the limits of human hand morphology.  
In this paper, we introduce a suite of teleoperated tasks that are challenging or impossible for the unaided human hand, thereby showcasing the unique advantages of our non-anthropomorphic hardware.

\textbf{Adhesive end-effectors} are those end-effectors that can apply force towards the contact surface from the object to the end-effector, and this has been concluded as astrictive prehension \cite{monkman1997analysis}.
Suction cups are widely used for robotic grasping with the negative fluid pressure \cite{mahler2018dex}. 
Other adhesive end-effectors include gecko inspired two-fingered gripper \cite{ruotolo2021grasping}, active adhesive materials on a five-fingered hand \cite{linghu2025versatile}, and three fingered submarine gripper based on water flow \cite{stuart2018tunable}. 
Suction cups on multi-fingered hands are not new for grasping, including a soft four-fingered hand with suction cups on the back of fingertips \cite{liu2023hybrid} and multi-fingered hands with suction cups on fingertips \cite{liu2022three, yamaguchi2013development}.
However, in these systems, adhesion is primarily used as a grasping aid to secure an object. 
The fundamental distinction of our SLeap Hand is its use of suction as an active tool for enabling complex in-hand manipulation, moving far beyond static grasping.

\textbf{Teleoperation} of robotic manipulation is crucial to collect real-world training data of high-quality. 
ALOHA, the dual arm system, has explored how to efficiently build a dual arm system to collect the training data directly in the real world \cite{aldaco2024aloha}.
However, this directly motion mapping strategy becomes extremely difficult for dexterous manipulation with multi-fingered hands, and several works were proposed to solve this problem.
DexGen used a pre-trained policy to understand human's intention during teleoperation \cite{yin2025dexteritygen}. 
DexForce used human-dragging to register the force during deployment and dragging \cite{chen2025dexforce}.
DoGlove proposed a new hardware that can be worn to feel the haptic feedback \cite{zhang2025doglove}.
A similar exoskeleton named DexOP is used to collect several in-hand manipulations \cite{fang2025dexop}.
However, these teleoperation systems either require pre-trained models on specific in-hand manipulation tasks, or complex re-targeting from human hand motion to robot hand motions at the hardware level.
Our SLeap Hand system where the hardware embodiment itself simplifies the teleoperation challenge. 
This allows our system to be independent of task-specific models and frees it from anthropomorphic constraints, thereby lowering the barrier for demonstrating super-human manipulation behaviors.

\section{SLeap Hand Design}
\label{sec:sys_design}
Our SLeap Hand design contains two parts: the hand design and the teleoperation system design.

\subsection{Hand Design}
\label{sec:hand_design}

\begin{figure}[t]
  \centering
  \includegraphics[width=0.48\textwidth]{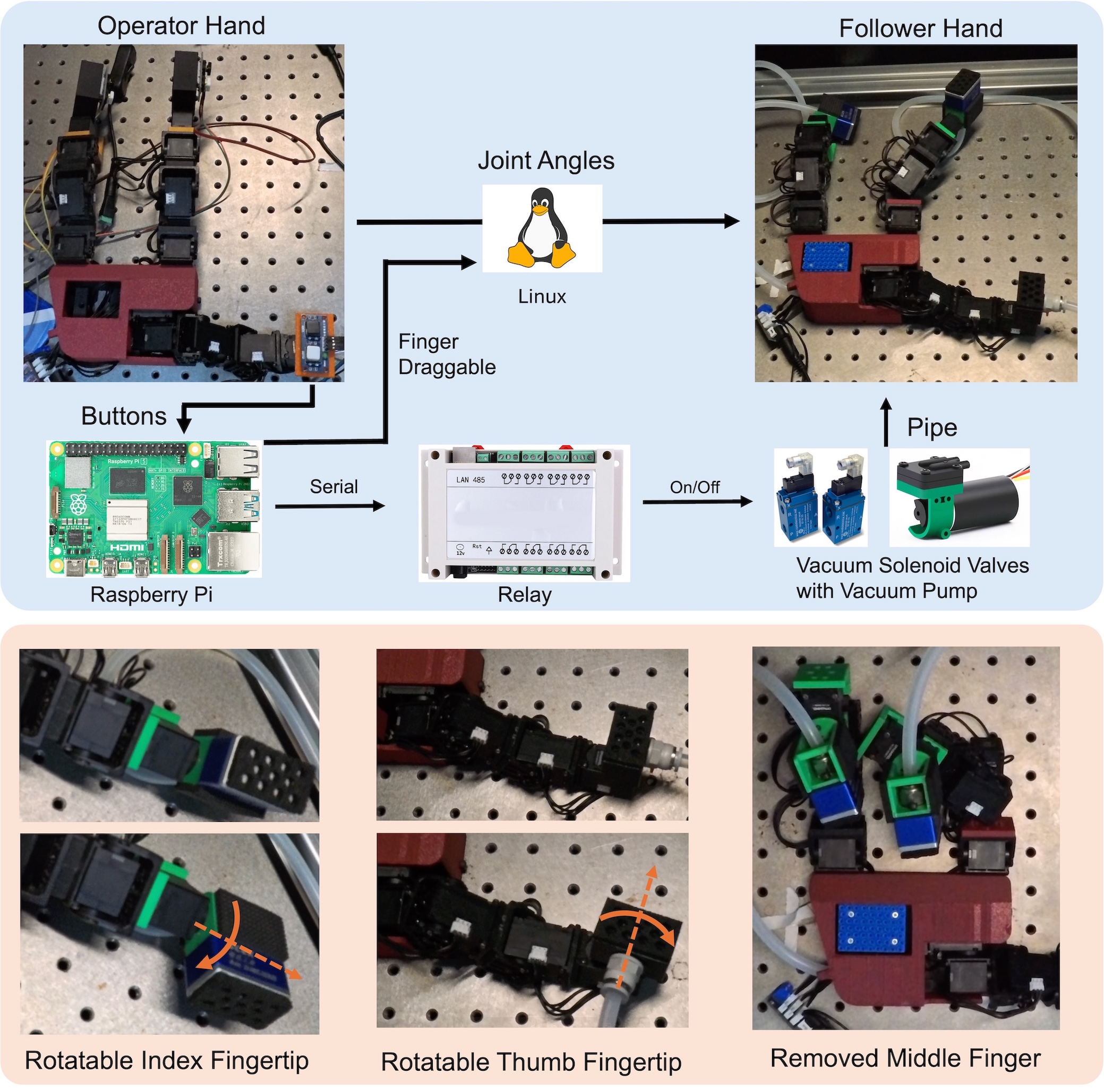}
  \caption{SLeap Hand Design. 
    \textbf{Top:} The operator hand transmits joint angles to the follower hand whenever the finger-drag buttons are pressed. 
    Suction cups are activated when the suction button is pressed, a relay triggers the vacuum solenoid valves to establish or release adhesion.
    \textbf{Bottom:} Left: Index and ring fingertips with the suction cups are rotatable, adding one extra joint. Middle: Thumb fingertip has a different rotation direction. Right: Middle finger is removed for more manipulation space of the index and the ring.
}
  \label{fig:design_hand}
  \vspace{-7mm}
\end{figure}

Following Leap Hand design \cite{shaw2023leap}, we keep the layout of the thumb, index, and ring fingers as well as their original design of four joints per finger, shown in Figure \ref{fig:design_hand}. 
The main differences are:

1. Each fingertip is equipped with a small suction cup. 
To enhance its adaptability to varied object geometries and materials, the rigid cup is covered with a layer of sponge, which ensures a reliable seal by improving surface compliance.

2. We further enhanced each finger with a fifth degree of freedom: axial rotation of the distal joint. This allows the fingertip to be precisely oriented, which is fundamental for achieving a flush contact between the suction cup and a target surface. To maximize manipulative dexterity, the thumb's rotational axis is orthogonal to the other fingers, enabling the hand to impart moments on an object for complex tasks like unscrewing the bottle cap.

3. We adopt a three-fingered configuration (thumb, index, ring), a strategic design choice that enhances manipulative capability without trade-offs. This streamlined layout provides a larger feasible space for fingers, which both reduces the likelihood of self-collision and creates more space for dexterous in-hand motions. Crucially, any potential loss in grasp stability from removing a finger is fully offset by the robust adhesion of the suction system. This allows the hand to maintain a secure grasp while benefiting from a simpler, more efficient mechanical design.

4. To leverage the palm as an active support surface, a suction cup is mounted on the palm. This introduces a distinct palmar grasping modality, enabling the hand to securely anchor objects without needing to engage the fingers. This feature is crucial for freeing the fingers to perform subsequent or simultaneous manipulation tasks on the stabilized object

\subsection{Teleoperation System Design}
\label{sec:teleop_design}

\begin{figure*}[t]
  \centering
  \includegraphics[width=0.98\textwidth]{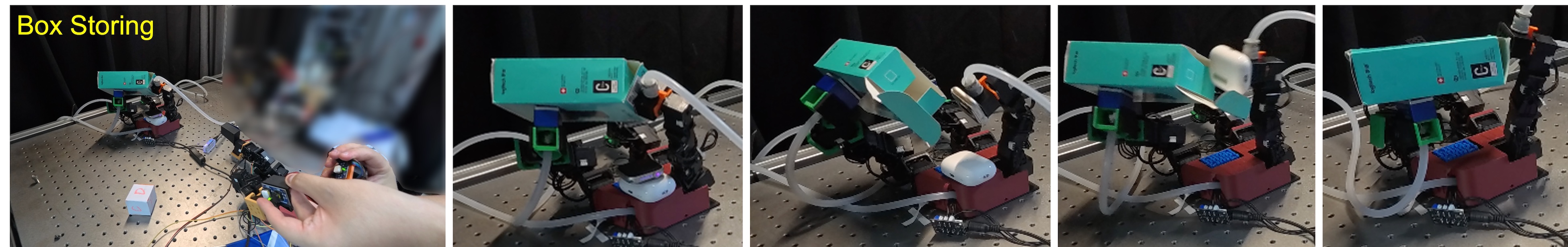}
  \caption{One case of using the teleoperation system for in-hand box opening and storage of a watch and an EarPods case.}
  \label{fig:design_teleop_demo}
  \vspace{-7mm}
\end{figure*}

Our teleoperation system is built on a leader-follower architecture inspired by ALOHA, consisting of an operator-side leader hand and a remote follower SLeap Hand. 
The follower kinematically mimics the leader's pose. 
To enable intuitive control, we have designed a custom interface on the leader hand. 
Each fingertip is equipped with two buttons: a white button that engages a gravity-compensated mode for effortless kinesthetic guiding of the finger, and a black button that toggles the activation of the corresponding suction cup on the follower hand.
This control interface is managed by a Raspberry Pi, which communicates with a PC via ROS2 and actuates the four vacuum solenoid valves through a relay. 
The complete teleoperation pipeline is illustrated in Figure \ref{fig:design_hand}, and a demonstration of the system performing a box-opening task is shown in Figure \ref{fig:design_teleop_demo}. A full breakdown of the hardware components and their associated costs is available in Table \ref{table:design_cost}

\begin{table}[h]
\centering
\caption{Cost to build a SLeap Hand teleoperation system. We assume the user already has a Leap Hand having 16 motors. Dynamixel Accessories include a powerhub, a USB communication converter, and several motor connectors. Adapters include one 5V adapter for all 30 motors and one 12V adapter for the vacuum pumps and vacuum valves.}
\label{table:design_cost}
\begin{tabular}{ccc}
\hline
Item                  & Num & Cost   \\ \hline
Dynamixel XC330-M288  & 15  & \$1500 \\
Dynamixel Accessories & 1   & \$200  \\
Suction Cup A + B        & 3 + 1   & \$60 + \$20   \\
Vacuum Pump (40L/min) & 1   & \$150  \\
Relay                 & 1   & \$30   \\
Valve                 & 4   & \$60   \\
Buttons               & 7   & \$10   \\
Adapter               & 2   & \$60   \\
3D PLA                & 1   & \$25   \\ \hline
Sum                   &     & \$2115 \\ \hline
\end{tabular}
\vspace{-3mm}
\end{table}

We compare our SLeap Hand teleoperation system with several previous methods on the dexterous demonstration collection, shown in Table \ref{table:comparsion}.
These methods either need pre-modeling or pre-training with Reinforcement Learning in the simulation for a specific task, or require direct human motion tracking, which constrains the robot hand to an anthropomorphic morphology.
Furthermore, few of these systems can perform robust in-hand reorientation in all three axes. 
In contrast, our SLeap Hand is not bound by human kinematics and can generate high-quality reorientation demonstrations without prior training, as we show in the expanded task suite in Section \ref{sec:experiment}.

\begin{table}[h]
\centering
\caption{Comparison with other demonstration collection methods. Requiring pre-modeled means the teleoperation requires either pre-training in simulation or pre-modeling for a specific task. Requiring human hand-like indicates the dexterous robotic hand limits to those human-hand like. Reorient indicates the hand can perform in-hand reorientation in all three axes. *Though our SLeap Hand has the thumb, the index, and the ring finger, there is no limitation on the specific structures since we do not require human hand motion as input.}
\label{table:comparsion}
\begin{tabular}{cccc}
\hline
Method                             & \begin{tabular}[c]{@{}c@{}}Required \\ Pre-modeled \end{tabular} & \begin{tabular}[c]{@{}c@{}}Required \\Human Hand-like \end{tabular} & Reorient \\ \hline
DexWild\cite{tao2025dexwild}       &      \xmark       &  \cmark               &   \xmark             \\
DexPilot\cite{handa2020dexpilot}   &      \cmark       &  \cmark               &   \xmark \\
DexOP\cite{fang2025dexop}          &      \xmark       &  \cmark               &   \cmark             \\
DOGlove\cite{zhang2025doglove}     &      \xmark       &  \cmark               &   \xmark             \\
DexGen\cite{yin2025dexteritygen}   &      \cmark       &  \cmark               &   \cmark             \\
DexForce\cite{chen2025dexforce}    &      \xmark       &  \xmark               &   \xmark             \\ 
AnyTeleop\cite{qin2023anyteleop}   &      \xmark       &  \cmark               &   \xmark \\ \hline

\textbf{SLeap Hand (Ours)}                         &     \xmark        &  \xmark*              &    \cmark            \\ 
\hline
\end{tabular}
\vspace{-5mm}
\end{table}

\section{Suction-based Embodiment}
Traditional dexterous hands are limited by contact-based mechanics: a stable grasp must rely on form-closure or force-closure conditions, where normal forces, friction, gravity, and object geometry are perfectly aligned to cage the object \cite{howard1996stability, bicchi1995closure}. 
This raises the control problem, which becomes extremely difficult to solve when a dexterous hand has many degree of freedoms.
In contrast, our work introduces active closure through suction-based adhesion, a form of astrictive prehension \cite{monkman1997analysis}.
Instead of passively depending on friction cones and opposing forces, suction provides a controllable adhesion force that effectively bypasses classical closure constraints. 
This shift yields three advantages for dexterous manipulation: 
1) it decouples the grasp from a specific joint configuration consisted of at least two fingers or even more, 
2) it enables inherently stable teleoperation, and 
3) it transforms complex in-hand reorientation into a series of simple grasps and grasping pose change, which are regrasping in the rest of the paper.

\subsection{Decoupling the Grasp from Constrained Joint Configuration}
\begin{figure}[h]
  \centering
  \includegraphics[width=0.48\textwidth]{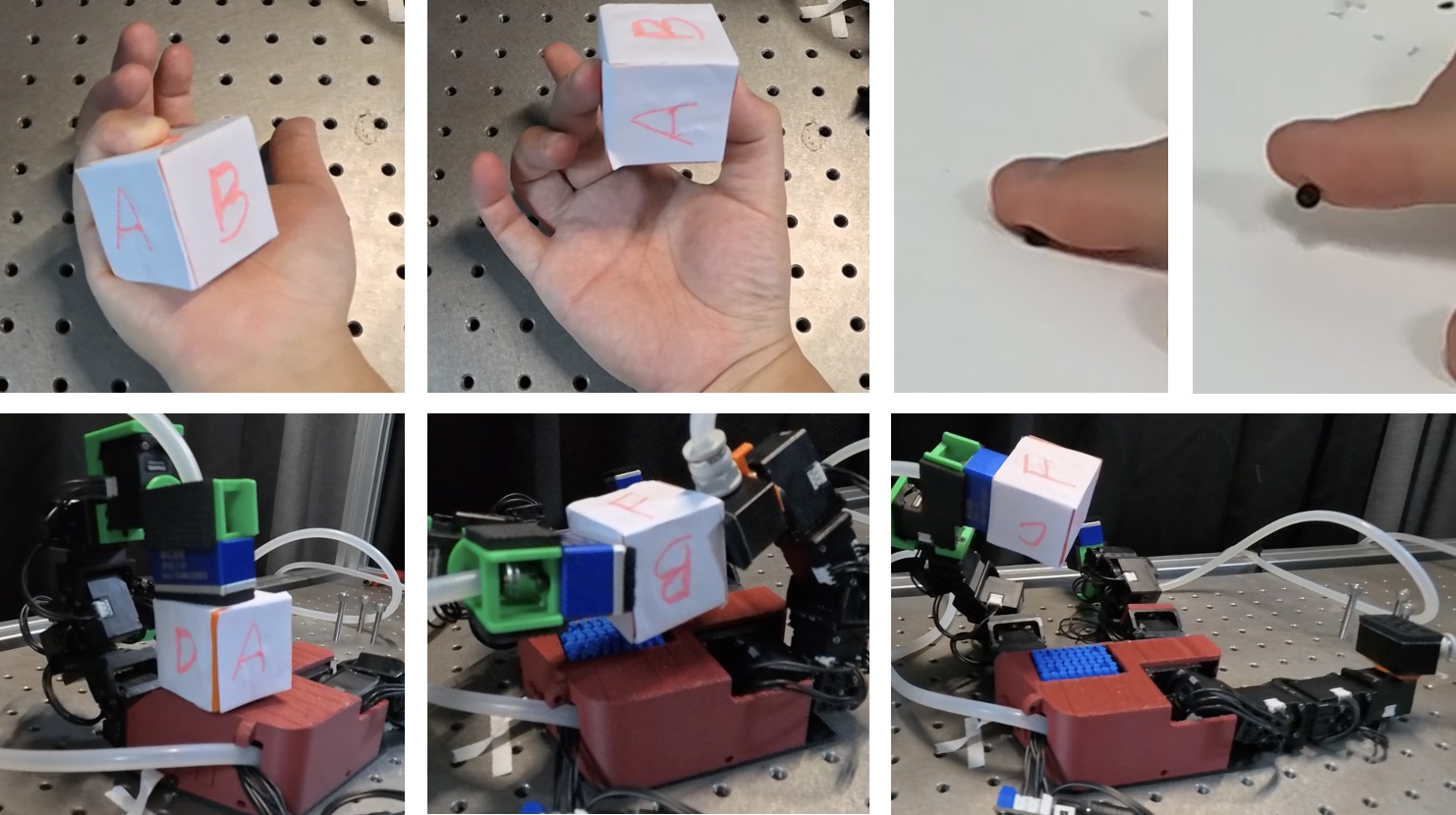}
  \caption{Contact-based grasping v.s. suction-based grasping from both a human hand and the SLeap Hand without using suction. Left: Finger-palm grasping. Middle: Finger-finger grasping. Right: Adhesive grasping, where human can also perform this to grasp objects of small size due to sticky skins \cite{spinner2016sticky}.}
  \label{fig:suc_non_constraint}
  \vspace{-6mm}
\end{figure}

A suction-based hand dramatically expands the space of available grasps. 
In a traditional force-closure grasp, the hand's joint configuration is heavily constrained  \cite{feix2015grasp}; 
fingers must be positioned in opposition to create a force cage, severely limiting the available contact points on an object's surface.
Finger-palm grasping loses the ability to change the object's pose, shown in Figure \ref{fig:suc_non_constraint} left.
Finger-finger grasping requires at least two fingers, and the object can only be reoriented in a small range, shown in Figure \ref{fig:suc_non_constraint} middle. 
In these two cases, traditional contact-based grasping limits the dexterity.

However, suction-based grasping removes this geometric restriction. 
A stable grasp can be achieved with a single point of contact as long as there is a proper smooth surface, shown in Figure \ref{fig:suc_non_constraint} right. 
This fundamentally changes the grasping problem from a complex, multi-finger geometric planning task to a simple search for a viable adhesion point. 
Consequently, the stability of the grasp is no longer coupled to the coordinated configuration of all fingers. 
One finger can secure the object with adhesion, leaving the remaining fingers free to rest, reposition for a future action, or provide auxiliary support, thus vastly enlarging the effective joint space for manipulation and keeping the redundancy.

\subsection{Stable Grasping}
The reliance on force-closure is a main reason why dexterous teleoperation remains a formidable challenge. 
Human operators, lacking direct haptic feedback, are often uncertain if a grasp is stable, as they must mentally model and maintain a delicate balance of unseen and unfelt contact forces. 
This cognitive load makes any subsequent in-hand manipulation a high-risk, tentative process.
Such uncertainty becomes severe during long-horizon dexterous manipulation.

Suction-based grasping converts this ambiguous, force-sensitive control problem into a clear, binary state. 
A grasp is either stable (vacuum seal maintained) or it is not. 
This certainty provides the operator with unambiguous feedback and confidence in the grasp's integrity.
By anchoring the object to the hand with a single, reliable suction point, the operator is freed from the constant burden of maintaining grasp stability. 
Their focus can shift entirely from holding the object to manipulating it, making the entire teleoperation process more robust, intuitive, and accessible.

\subsection{Grasping-based Reorientation}
Beyond grasping, suction-based embodiment redefines the mechanics of in-hand reorientation.
With conventional hands, reorienting an object is a highly skilled process involving carefully coordinated finger gaiting, including a sequence of controlled rolling and sliding motions where fingers pass the object between one another. 
This process is dynamically complex and prone to failure if even one contact is misplaced.

\begin{figure}[h]
  \centering
  \includegraphics[width=0.48\textwidth]{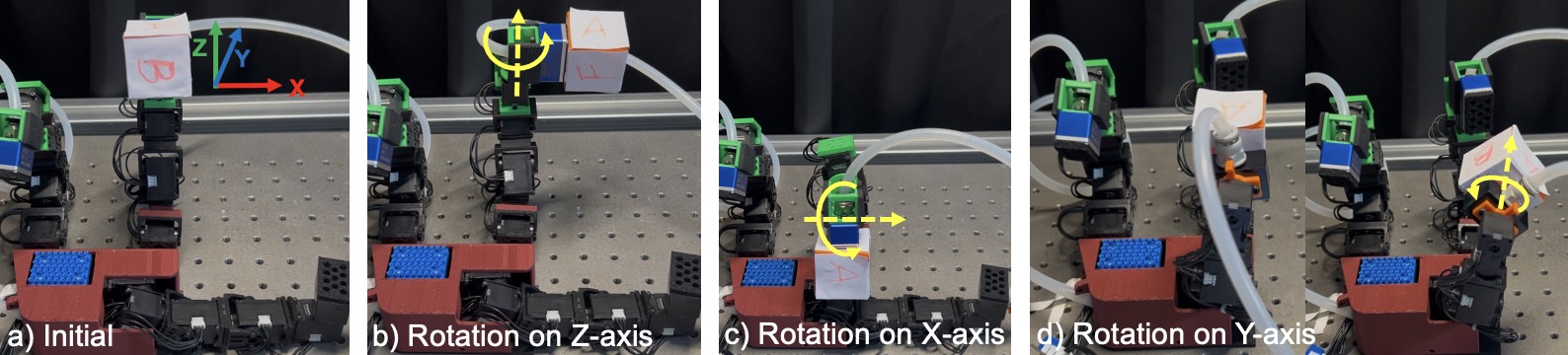}
  \caption{In-hand cube reorientation with rotatable fingertips and multiple regrasps to other fingers. Rotations in all three axes can be performed with suction-based grasping.}
  \label{fig:suc_reorient}
  \vspace{-5mm}
\end{figure}

Suction transforms this dynamic juggling act into a far simpler, quasi-static procedure built on sequential regrasping. 
Because adhesion provides a stable anchor point, an object can be firmly held by a single fingertip.
This single point of contact secures the object, liberating the other fingers to completely release their contact, reconfigure freely in space, and establish new grasp points on the object.
The complex problem of maintaining continuous, multi-point force-closure is replaced by a simple, discrete sequence: grasp, reconfigure, and regrasp, shown in Figure \ref{fig:suc_reorient}.

This allows the hand to effectively move its fingers across an object's surface, achieving large-angle rotations and translations that would be impossible with friction-based grasps, where the object must be continuously caged.
In effect, suction decomposes complex in-hand reorientation into a series of simple, stable grasp-and-release primitives. 
This not only dramatically simplifies the control problem but also unlocks a vast new space of manipulation trajectories unavailable to traditional anthropomorphic hands.

\section{Experiment}
\label{sec:experiment}
In this section, we provide empirical validation by teleoperating the hand through a suite of complex tasks, including several that are considered intractable for conventional multi-fingered hands.

\begin{figure*}[t]
  \centering
  \includegraphics[width=1\textwidth]{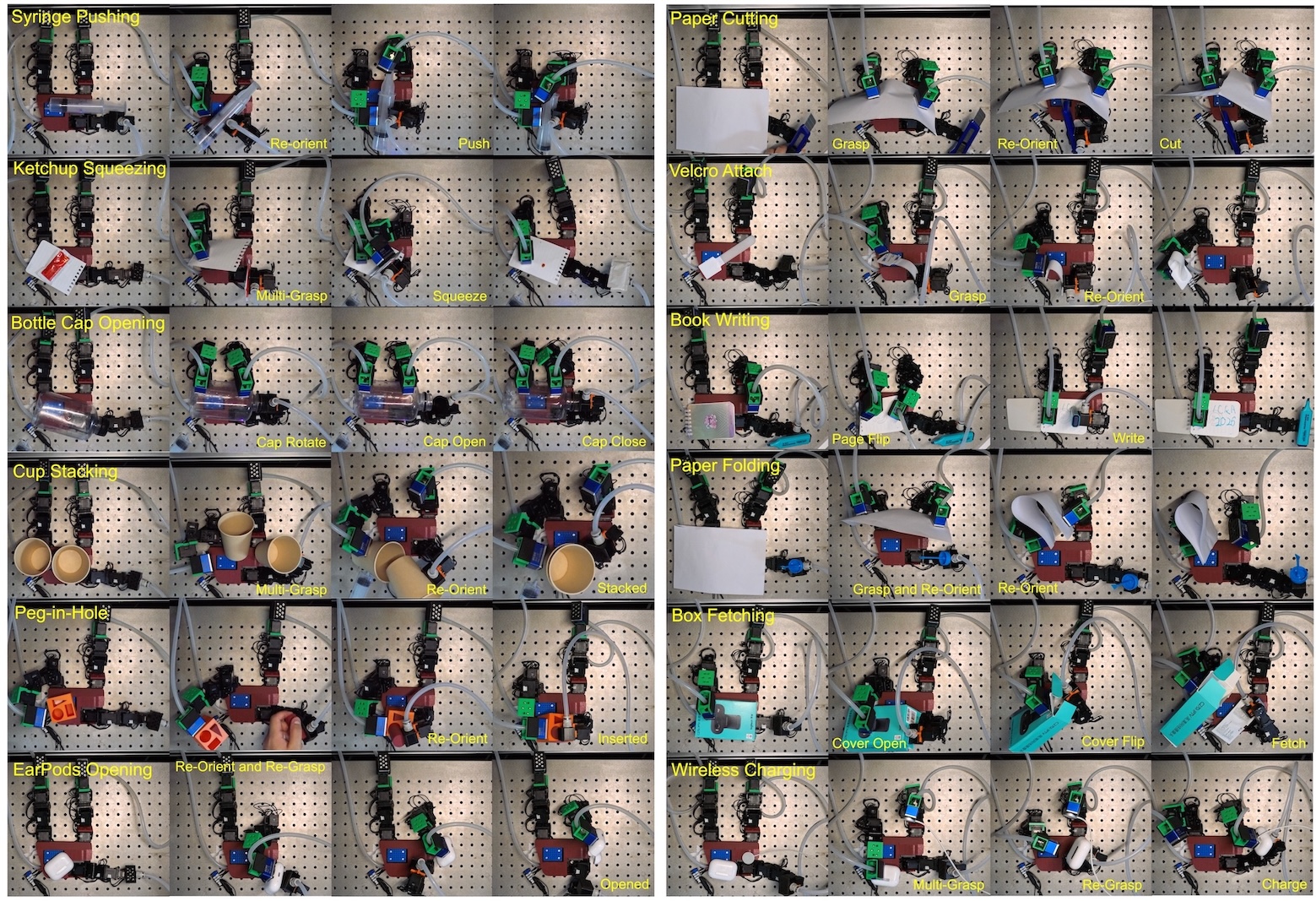}
  \caption{12 challenging in-hand manipulation tasks performed by teleoperations. Key steps are highlighted in the figure.}
  \label{fig:exp_teleop}
  \vspace{-5mm}
\end{figure*}

\begin{figure}[h]
  \centering
  \includegraphics[width=0.48\textwidth]{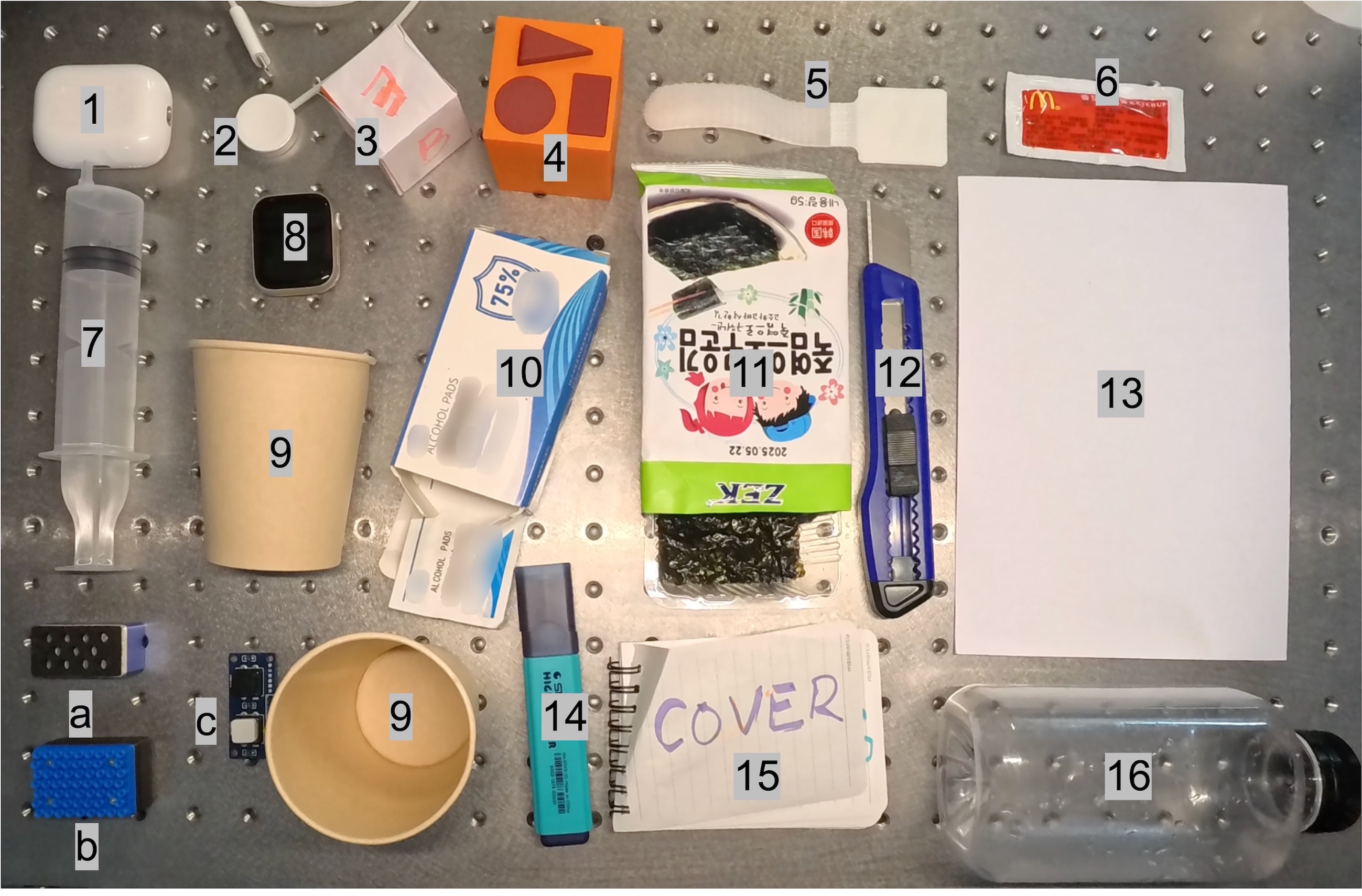}
  \caption{All objects that are used in the experiments. a) Suction cup A on fingertips. b) Suction cup B on the palm. c) Two buttons on the operator hand. 1) EarPods case. 2) Wireless charger. 3) Cube with A to F on 6 faces. 4) Pegs and a hole object. 5) Velcro strip. 6) Ketchup packet. 7) Syringe. 8) Watch. 9) Cup. 10), 11) Box and the object inside. 12) Knife. 13) A5 paper. 14) Marker pen. 15) Notebook. 16) Bottle with cap.}
  \label{fig:exp_all_obj}
  \vspace{-8mm}
\end{figure}

\subsection{Task Definitions}
We define 14 common tasks, including long-horizon tasks and those performed by previous teleoperation methods, and some are even challenging for human, shown in Figure \ref{fig:exp_teleop}:

1. Syringe Pushing \cite{yin2025dexteritygen}: A syringe is placed on the palm. The hand is required to grasp and reorient the syringe into a functional pushing pose, and then depress the plunger. This task is broken down into two primary steps:
a. Reorientation: Grasping and rotating the syringe to the correct pose for operation.
b. Pushing: Depressing the syringe plunger.

2. A single-serving ketchup packet and a small piece of paper are placed on the palm. The task is to squeeze a portion of ketchup onto the paper. Successful completion is achieved when ketchup is transferred to the paper. The task involves two steps:
a. Grasping: Securing both the paper and the ketchup packet.
b. Squeezing and Dipping: Manipulating the packet to dispense ketchup onto the paper.

3. Bottle Cap Opening: An empty bottle with a screw-cap is placed on the palm. The hand must first grasp the bottle and unscrew the cap until it is fully detached. The core steps for evaluation are:
a. Grasping and Reorientation: Positioning the bottle so the thumb and fingers can effectively manipulate the cap.
b. Unscrewing: Rotating the cap until it is separated from the bottle.

4. Earpods Opening \cite{chen2025dexforce}: An Apple Earpods case is placed on the palm. The hand must grasp and reorient the case to an appropriate position, and then use its fingers to open the lid. The task consists of two steps:
a. Grasping and Reorientation: Stabilizing the Earpods case in a suitable pose.
b. Lid Opening: Manipulating the lid to open the case.

5. Cube Rotation: A cube is placed on the palm, and the hand must rotate it three full times around each of its three primary axes (X, Y, and Z) in any order. The task is segmented into three steps:
a, b, c. Rotation on X, Y, and Z-axis: Three successive rotations about the X, Y, and Z-axis.

6. Cup Stacking: Two cups are placed on the palm. The objective is to grasp both cups and stack one inside the other. This task is performed in three steps:
a. First Cup Grasp and Reorientation: Grasping and maneuvering the first cup.
b. Second Cup Grasp and Reorientation: Grasping and maneuvering the second cup.
c. Stacking: Inserting one cup into the other.

7. Peg-in-Hole: A block containing three distinct holes (triangular, rectangular, cylindrical) is placed in the hand. Corresponding pegs (a triangular prism, a rectangular cuboid, and a cylinder) are sequentially delivered to the hand by a human operator. The hand must insert each peg into its matching hole. The four steps are:
a. Grasping Block: Securing the object with the holes.
b, c, d. Peg Insertion: Performing the three peg-in-hole insertions.

8. Paper Folding: An A5-sized piece of paper is placed in the hand. The task is to grasp the paper from two sides and create a fold. The process involves two steps:
a. Grasping: Securing two sides of the paper.
b. Folding: Bringing the grasped sides together to create a crease.

9. Paper Cutting: The hand first grasps an A5-sized piece of paper. A utility knife is delivered by an operator. The hand must then use the knife to pierce the paper in the middle. The two main steps are:
a. Paper Grasping: Stabilizing the piece of paper.
b. Piercing: Using the provided knife to cut through the paper.

10. Book Flipping and Writing:  A small notebook is placed in the hand. The hand must first open the cover and then, using a marker pen delivered by an operator, write \textit{ICRA 2026}. The task is divided into three steps:
a. Flipping Cover: Opening the notebook to the first page.
b. Writing \textit{ICRA}: Inscribing the first word on the page.
c. Writing \textit{2026}: Inscribing the second word.

11. Box Opening and Object Retrieval: A box is placed on the palm. The hand needs to open the box and retrieve an object from within. The task consists of three steps:
a. Grasping Box: Securing the box in-hand.
b. Opening Flap: Manipulating and opening the side flap of the box.
c. Object Retrieval: Grasping and removing the object from the box.

12. Hook-and-Loop Fastener (Velcro) Attachment: The two parts of a Velcro strip are presented to the hand. The task is to join the two sides to fasten them. This involves two steps:
a. Grasping Strips: Holding both sides of the fastener.
b. Attachment: Pressing the two sides together to engage the hook-and-loop mechanism.

13. Box Opening and Object Storing: A box is placed on the palm. An object is delivered by an operator, and the hand must place this object inside the box and then close it. The three steps are:
a. Grasping Box: Stabilizing the box in the hand.
b. Opening Flap: Opening the side flap of the box.
c. Storing Object: Placing the delivered object inside the box.

14. Wireless charging: The hand is required to charge an Apple Watch using a wireless charger. The task is broken down into two steps:
a. Watch Grasping and Reorientation: Grasping the watch and positioning it with its back exposed.
b. Charger Alignment: Aligning the charger with the back of the watch to initiate charging.

\subsubsection{Environment Setup}
Both the operator Hand and the follower Hand are fixed with palm facing upward on an optical platform, shown in Figure \ref{fig:design_teleop_demo}.
All objects that are used in the experiments are shown in Figure \ref{fig:exp_all_obj}.

\subsection{Evaluation on Dexterous Teleoperation}

\begin{table*}[thbp]
\centering
\caption{Task Definition and Evaluation. M indicates that the task involves \textbf{M}ultiple object manipulation, D is \textbf{D}eformable object manipulation,  A is \textbf{A}rticulated object manipulation, R is \textbf{R}e-orientation. The second part is the evaluation of our teleoperation system.  For human hand comparability, we have \cmark \space for tasks that are achievable by a human hand, regardless of the initial object pose on the palm,
\xmark \space for tasks that are considered nearly impossible for a single human hand to perform, and 
\cmark \xmark \space for tasks that are performable by a human hand only if the objects are placed in a favorable initial configuration.}
\label{table:exp_task_evaluation}
\centering
\begin{tabular}{c|ccccc|cccc}
\hline
Task                                                                     & M                     & D                     & A                     & R                     & Steps & Overall SR & Step Completeness & Ave Time (s) & Human Hand Comparability \\ \hline
Syringe Pushing                                                          & \xmark & \xmark & \cmark & \cmark & 2     &   0.7         &   0.85      &    43      &                   \cmark           \\
Ketchup Squeezing \& Dipping                                             & \xmark & \cmark & \xmark & \cmark & 2     &   0.8         &   0.9      &    84      &      \cmark         \\
Bottle Cap Opening  & \cmark & \xmark & \cmark & \cmark & 2     &   0.2         &   0.5      &    118      &      \cmark             \\
EarPods Opening                                                          & \xmark & \xmark & \cmark & \cmark & 2     &    0.6        &  0.75       &   47       &  \cmark             \\
Cube Rotation               & \xmark & \xmark & \xmark & \cmark & 3     &   0.6         &    0.7     &   84       &             \cmark      \\
Cup Stacking                                                             & \cmark & \xmark & \xmark & \cmark & 3     &    0.8        &   0.93      &   28       &              \cmark \xmark              \\
Peg-in-Hole                                                              & \cmark & \xmark & \xmark & \cmark & 4     &    0.5        &   0.75      &  195        &              \cmark \xmark                 \\
Paper Folding                                                            & \xmark & \cmark & \xmark & \cmark & 2     &    0.6        &   0.8      &  62        &              \cmark               \\
Paper Cutting                                                            & \cmark & \cmark & \xmark & \cmark & 2     &    0.7        &   0.85      &   50       &              \xmark               \\
Book Flipping \& Writing    & \cmark & \xmark & \cmark & \cmark & 3     &   0.8         &   0.93      &   157       &              \xmark                 \\
Box Opening   \& Fetching      & \cmark & \xmark & \cmark & \cmark & 3     &   0.5         &  0.67       &  198        &              \cmark \xmark            \\
Velcro Attach                                                             & \xmark & \cmark & \xmark & \cmark & 2     &    0.8        &  0.9       &   21       &              \cmark                \\
Box Opening   \& Storing       & \cmark & \xmark & \cmark & \cmark & 2     &   0.5         &   0.65      &   144       &              \cmark \xmark               \\
Wireless Charging                                                        & \cmark & \cmark & \cmark & \cmark & 2     &    1.0        &   1.0      &    35      &              \cmark              \\ \hline
\end{tabular}
\vspace{-5mm}
\end{table*}

Each task was attempted 10 times to ensure robust evaluation. 
For most tasks, all objects were initially placed on the palm of the hand in various poses to test adaptability.
However, for the peg-in-hole, book flipping and writing, and paper cutting tasks, the respective tool (peg, pen, or knife) was pre-positioned on the thumb fingertip manually.

Metrics for Evaluation:
We assessed the system's usability and performance using the following metrics:

1. Overall Success Rate: The percentage of trials completed without any critical failures like dropping.

2. Step Completeness: The average percentage of completed steps across all trials for a given task. A fully successful trial contributes 100\% to this metric. If a task with four steps fails after completing two, it contributes 50\%.

3. Average Time: The mean time elapsed from the initiation of the task to its successful completion.

4. Human Hand Comparability: A qualitative assessment of whether a task can be performed by a single human hand.

\subsection{Results}
The experimental results are summarized in Table \ref{table:exp_task_evaluation}.
These tasks all require at least two steps with mixed multiple object manipulation, deformable object manipulation, and articulated object manipulation.
All tasks require in-hand reorientation.
The SLeap Hand teleoperation system demonstrated strong performance across these challenging, long-horizon tasks.

\begin{figure}[h]
  \centering
  \includegraphics[width=0.48\textwidth]{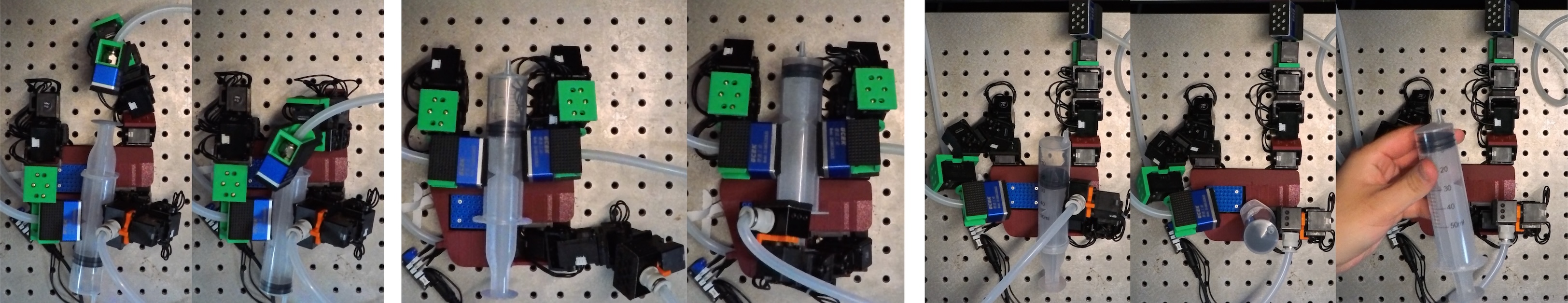}
  \caption{Versatile strategies for pushing a syringe. Left: The syringe is grasped by the thumb and ring finger and pushed by the index finger. Middle: The syringe is grasped by the index and ring fingers and pushed by the thumb. Right: The syringe is held and pushed using only a single finger, facilitated by suction.}
  \label{fig:exp_versatile}
  \vspace{-5mm}
\end{figure}

Furthermore, the system exhibited versatility in its manipulation strategies for the same task. 
As shown in Figure \ref{fig:exp_versatile}, the syringe-pushing task was completed in three distinct ways. 
The third method, where the syringe is held and actuated by a single finger, is a unique capability enabled by the suction-based grasping mechanism.

\subsection{Analysis of Failure Cases}
\begin{figure}[h]
  \centering
  \includegraphics[width=0.48\textwidth]{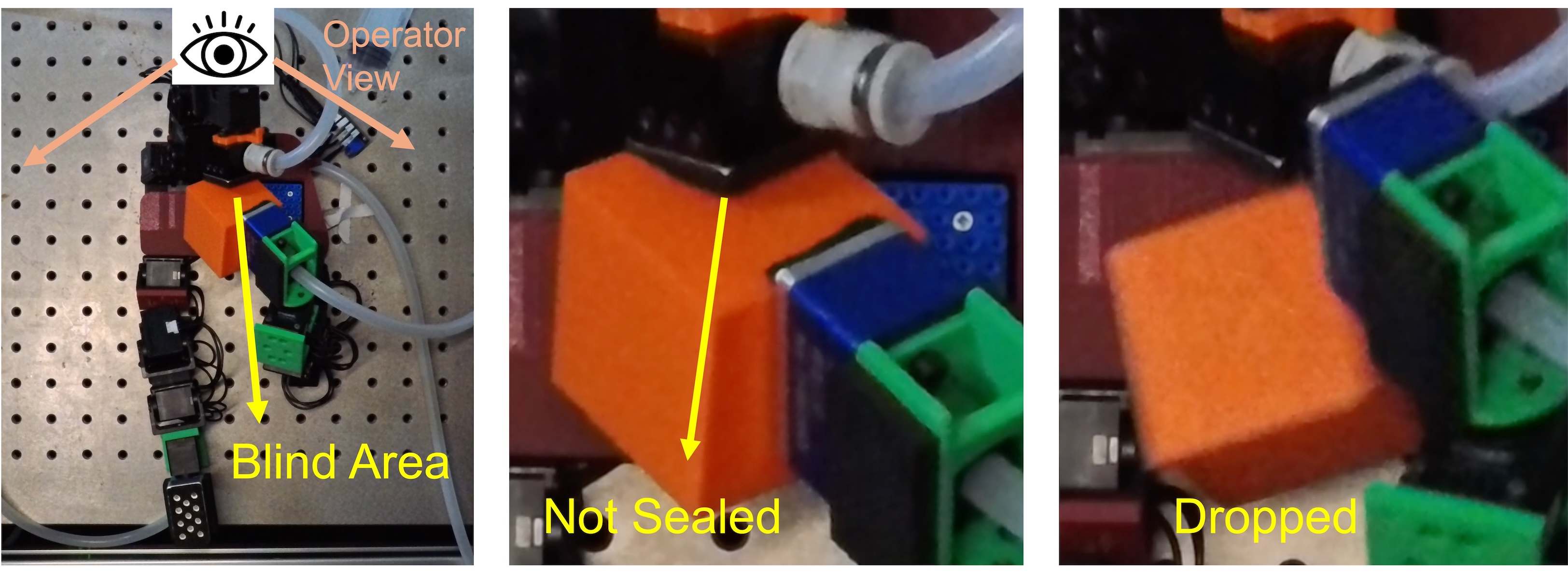}
  \caption{Failure due to blind area from the operator's sight.}
  \label{fig:exp_failure}
  \vspace{-6mm}
\end{figure}

The most common failure mode observed during teleoperation was object dropping. 
While the integration of suction significantly improves grasp stability, failures still occurred. 
The primary cause was linked to visual occlusion, shown in Figure \ref{fig:exp_failure}. 
The operator could not always accurately perceive the contact and seal quality of the suction cup, particularly on surfaces hidden from view. 
This could lead to an incomplete seal, causing the object to slip or be dropped during subsequent regrasping.

\section{Limitations}
We identify several limitations of the current SLeap Hand:

1. Lack of Arm and Wrist Integration: 
    The current research focuses exclusively on in-hand manipulation, without integrating arm or wrist movements.
    This prevents the realization of more complex, whole-body synergies between the hand and the arm. 
    Future work will involve mounting the hand on a robotic arm with gravity compensation, allowing the operator to guide the arm and hand in a coordinated manner.
   
2. Constraints of Suction-Based Grasping:  
   Suction grasping is not universally applicable. 
   It is ineffective for very small objects that lack a flat surface area sufficient for a seal, as well as for deformable or porous materials like cloth. 
   Other objects, including round objects, may be difficult to secure and can roll off the fingertip when they are heavy. 
   The maximum suction force is approximately 6.5N on a flat surface and decreases when applied to curved or non-ideal surfaces.
   Although the hand can still use traditional contact-based grasping in scenarios where suction-based grasps fail, the teleoperation becomes more challenging and less reliable.

\section{Conclusion}
In this work, we presented the SLeap Hand, a novel three-fingered manipulator featuring suction cups on each rotatable fingertip and the palm. 
This design significantly enhances capabilities for performing complex tasks through both direct control and teleoperation. 
The core innovation lies in the integration of a suction-based embodiment into a dexterous hand. 
This allows individual fingers to be decoupled for independent grasping, which in turn leads to greater stability during teleoperation.
Consequently, challenging long-horizon tasks can be simplified into a sequence of reliable regrasps. 
Our teleoperation experiments demonstrated that this system enables human operators to successfully perform a suite of difficult in-hand manipulation tasks, including some that are exceptionally challenging for an unassisted human hand. 
Finally, the SLeap Hand teleoperation system serves as an effective platform for collecting high-quality dexterous manipulation demonstrations.
Crucially, because the hand can perform tasks that are challenging or impossible for a human, the collected data captures not only expert motions but also novel, super-human manipulation strategies. 
We believe these unique datasets will be an invaluable resource for training data-driven policies that can discover and leverage non-anthropomorphic solutions, accelerating progress in the field of robotic manipulation.

\bibliographystyle{IEEEtran}
\bibliography{scibib}

\end{document}